\documentclass[letterpaper, 10 pt, conference]{ieeeconf}  

\IEEEoverridecommandlockouts                              

\usepackage{graphics} 
\usepackage{epsfig} 
\usepackage{cite}
\usepackage{color}
\usepackage{booktabs}  %
\usepackage{amsmath}
\usepackage{amssymb}
\usepackage{subfigure}
\usepackage{indentfirst}
\usepackage[colorlinks,linkcolor=blue]{hyperref}

\title{\LARGE \bf
Learning Actions from Human Demonstration Video for Robotic Manipulation
}

\author{Shuo Yang, Wei Zhang${^*}$, Weizhi Lu, Hesheng Wang, and Yibin Li 
\thanks{S. Yang, W. Zhang, W. Lu and Y. Li are with the School of Control Science and Engineering, Shandong University, China.  H. Wang is with the Department of Automation, Shanghai Jiao Tong University, China. \rule[0.20\baselineskip]{3cm}{0.5pt}\newline
 \hspace*{0.25cm}${^*}$Corresponding author} 
}

\begin{document}

\maketitle
\thispagestyle{empty}
\pagestyle{empty}

\begin{abstract}

Learning actions from human demonstration is an emerging trend for designing intelligent robotic systems, which can be referred as video to command. The performance of such approach highly relies on the quality of video captioning. However, the general video captioning methods focus more on the understanding of the full frame, lacking of consideration on the specific object of interests in robotic manipulations. We propose a novel deep model to learn actions from human demonstration video for robotic manipulation. It consists of two deep networks, grasp detection network (GNet) and video captioning network (CNet). GNet performs two functions: providing grasp solutions and extracting the local features for the object of interests in robotic manipulation. CNet outputs the captioning results by fusing the features of both full frames and local objects. Experimental results on UR5 robotic arm show that our method could produce more accurate command from video demonstration than state-of-the-art work, thereby leading to more robust grasping performance.


\end{abstract}

\section{INTRODUCTION}



Learning actions from human demonstrations is an attractive capability for  robotic systems, which enables to learn more complex skills and tasks without the need of manually programming. Currently, this technique remains challenging due to the difficulty of understanding human actions.  As robots become more and more omnipresent, there is an increasing need for developing intelligent robotic systems which can understand human demonstrations and perform diverse tasks. 

An intuitive method for imitating human actions is to record the human body motion trajectories by using wearable sensors. For instance, Koenemann et al. \cite{4} captured the human motions using inertial sensors attached to the individual body segments. Also, one can directly teach the robots to carry out a task by physically moving it through a desired trajectories, such as the known kinesthetic teaching \cite{1}. Calinon et al. \cite{3} propose to make the robot learn the  actions of itself generated by the remote control of human.  Although these methods work well for specific tasks, the diversity of actions they could learn is limited due to involving extra physical systems to assist learning. 




The pioneering video to command (V2C) attempt was made by Nguyen et al. \cite{5}, which shows that the robotic actions can be learned directly from human demonstration videos. Given a human demonstration video, they first extract visual features, and then feed them into an encoder-decoder architecture to generate a command sentence. The command is finally executed by the robot using an affordance detection network \cite{do2018affordancenet}. By this procedure, the problem of understanding human actions is simply cast as a video captioning task which describes visual contents with natural language. So the major advantage of \cite{5} lies in the adoption of a general video captioning approach \cite{venugopalan2014translating} \cite{venugopalan2015sequence}. Since the general video captioning approach  focuses on the global feature of the full frame, rather than  the local features of manipulated objects, the method in \cite{5} tends to perform worse in complex scenarios, such as the pick-and-place tasks shown in our experiments.




\begin{figure}[t]
\begin{center}
\includegraphics[width=0.9\linewidth]{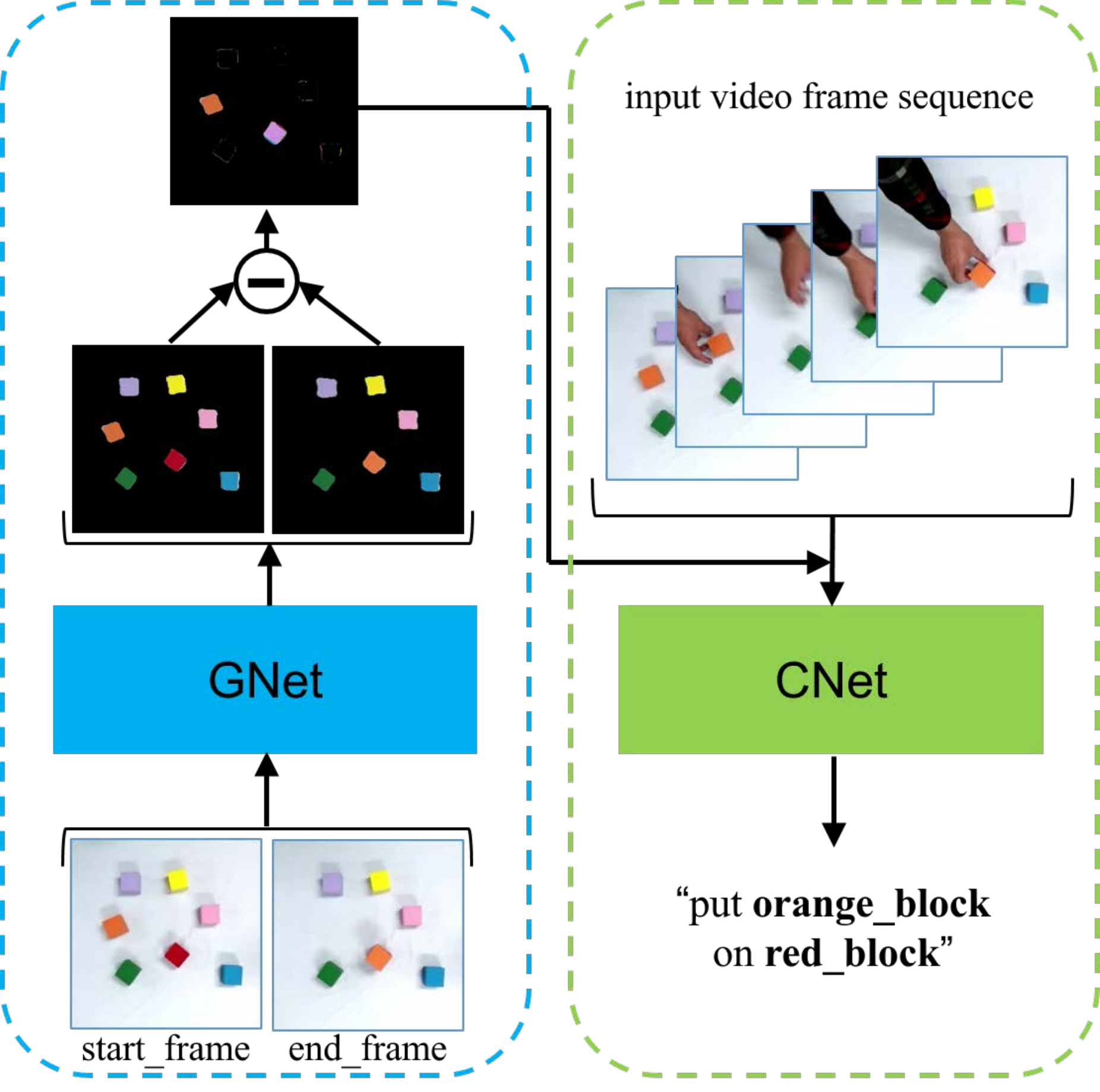}
\end{center}
\caption{Overview of our approach.}
\label{fig1}
\end{figure}

To make the video captioning  more adaptable to our robotic manipulation tasks, we propose to  fuse the global features (full frame) and local features (manipulated objects) together. For this purpose, we design a robotic system consisting of  two modules, grasp detection network (GNet) and video captioning network (CNet). (see Fig.~\ref{fig1}). Similar to \cite{5}, we use a CNet to  translate human demonstration videos to command sentences. The difference is that our CNet takes as input not only global features but local features. The local features come from a specially designed GNet, which mainly serves to provide grasp solutions. Note that  in \cite{5} the grasp solutions are generated by an affordance detection network \cite{do2018affordancenet}, which however cannot provide the local features we desire and thus is not applied in our method. In experiments, our method presents desired grasping performance by obtaining better captioning results than \cite{5}. 


The remainder of this paper is organized as follows. We provide an overview of the related work in Section II, then describe the proposed method in Section III. Experiments are presented and discussed  in Section IV. Finally, the paper is concluded  in Section V.

\section{RELATED WORK}

Robots with the capability of learning  actions from human demonstrations have been widely studied in the robotics community. In this section, we review the work mostly related to our approach.

Most of the methods of learning  human actions  operate at the level of configuration-space trajectories \cite{7} \cite{6} \cite{travaglini2016initial}. Some researchers used kinesthetic teaching as the approach to provide demonstrations for a robot to learn from demonstrations \cite{1} \cite{8} \cite{steinmetz2015simultaneous}. However, it is difficult to collect suitable demonstrations for real-world robotic manipulation via kinesthetic teaching. Another approach is to learn tasks through demonstrations which were captured by teleoperation devices \cite{10} \cite{11} \cite{yang2017teleoperation}. The work in \cite{12} \cite{13} \cite{peters2016feasibility} tried to capture the human motions with a motion capture system consisting of sensors which were installed on the demonstrator. Rojas et al. \cite{rojas2017motion} used a Xbox to collect data for the evaluation of people with hip disease. Li et al. \cite{brigante2011towards} \cite{wu2016wearable} developed a wearable system for real-time human motion capture to assist learning. Although these methods successfully worked, it is costly to capture the training demonstration data for the need of extra expensive physical systems, such as virtual reality headsets and motion capture sensors.

Intuitively, imitating human actions just by watching the demonstration videos is more meaningful and practical in real-world robotic applications. Welschehold et al. \cite{17} proposed to transform human demonstrations to different hand-object trajectories in order to adapt to robotic manipulation tasks. The work in \cite{16} \cite{moon2016multiple} \cite{farooq2015dense} studied the human pose tracking task by estimating human pose and human parts positions in demonstration videos. Aksoy et al. \cite{18} introduced a framework that represents the continuous human actions as semantic event chains (SEC) and identifies the types of performing manipulations according to SEC. However, these methods mostly need explicit representations (e.g. objects, hand grasping types, and human pose) and do not provide further meaningful clues that can be used in robotic applications.

Recently, Nguyen et al. \cite{5} proposed to learn actions from human demonstration videos by casting the the problem of understanding human actions as a video captioning task. The caption is translated to a command for robotic actions. This learning process seems more intuitive and practical, and  probably enables the robotic system  to learn complex skills or tasks without the need of tedious manual programming. Therefore, the performance of such method highly relies on the quality of video captioning. However, a general captioning approach \cite{venugopalan2014translating} \cite{venugopalan2015sequence} is adopted in \cite{5}, which focuses on the understanding of the full frame, rather than the objects of interest in robotic applications. This limits their performance in complex scenarios, such as the pick-and-place tasks shown in our experiments.   To address this issue, we propose a novel framework to improve the performance of  video captioning and object grasping.



\section{PROPOSED METHOD}

We introduce a novel approach to improve the performance of learning actions from human demonstration videos for robotic applications. The approach is built on two models: grasp detection network (GNet)  and video captioning network (CNet). GNet is developed to provide grasp solutions for commands execution and to improve the performance of CNet by providing a video difference map. CNet is used to translate human demonstration videos to command sentences.  The two models are detailed as follows.

\begin{figure}[t]
\centering
\includegraphics[width=0.6\linewidth]{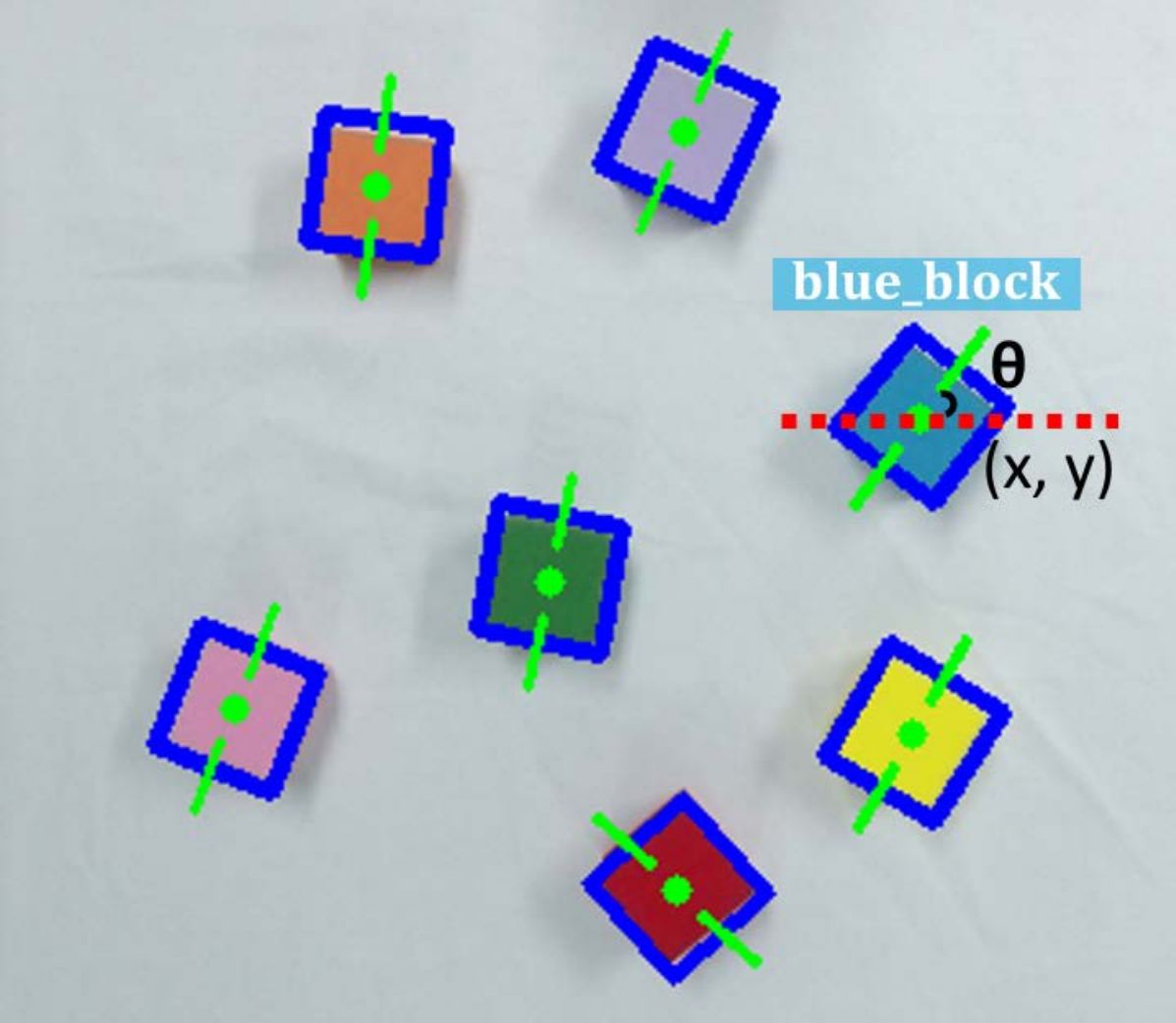}
\caption{A grasp representation, (x,y) corresponds to the center of grasp rectangle, \(\theta\) is the orientation of the rectangle relative to the horizontal axis, `blue\_block' is the object category label. \label{fig2}}
\end{figure}

\subsection{GNet based  Robotic Grasp Detection} In general, the robotic grasp detection aims to find a solution to pick up and hold the objects in a given image. 
Similarly to \cite{19} \cite{redmon2015real},  we define the representation of a grasp solution  as follow:

\begin{figure*}[t]
\begin{center}
\includegraphics[width=0.9\linewidth]{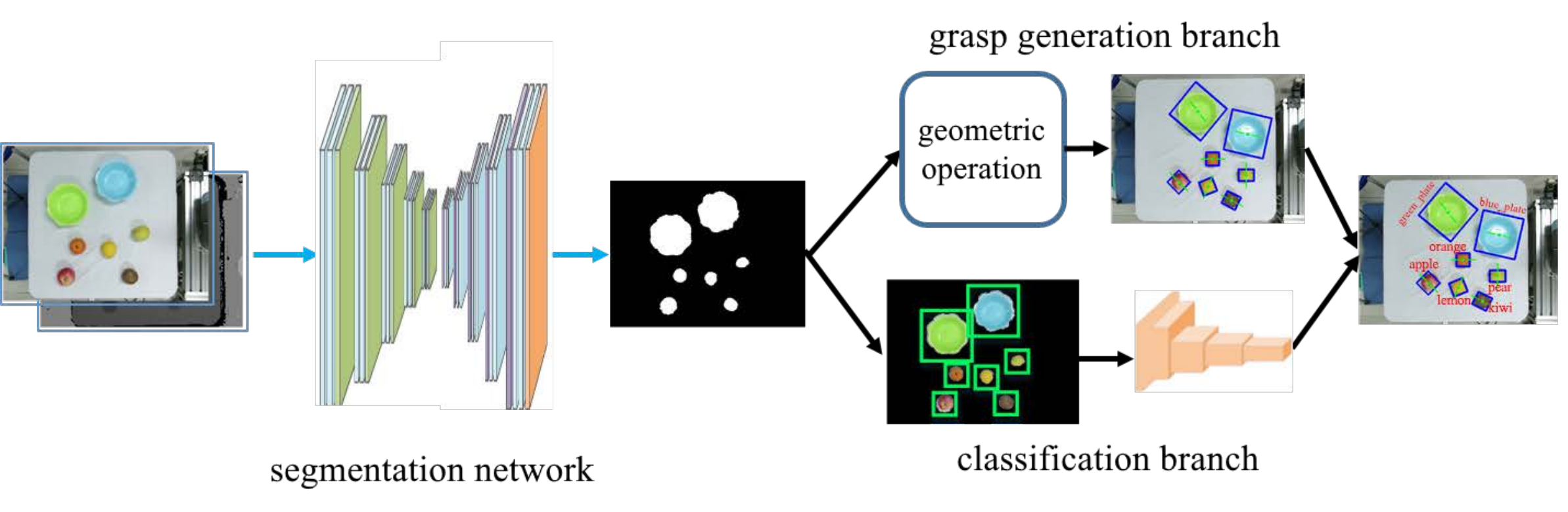}
\end{center}
\caption{An illustration of our GNet. \textbf{From left to right:} A segmentation network is first used to generate object mask. Then the grasp generation branch  calculates the grasp position and orientation of each object. The classification branch  generates a color mask and classify the object region through a CNN. Finally, the grasp position, orientation and category label are combined as a complete grasp representation.}
\label{fig3}
\end{figure*}

\begin{equation}
  \begin{split}
  g = \{x,y,\theta,l\},
  \end{split}
\end{equation}
where (\textit{x, y}) corresponds to the center of grasp rectangle, \(\theta\) is the orientation of the rectangle relative to the horizontal axis, and \textit{l} denotes the object category label. Fig.~\ref{fig2} shows an example of our grasp representation.

To efficiently detect the robotic grasp, we propose GNet, which combines object segmentation and object classification in an end-to-end manner. As shown in Fig.~\ref{fig3}, GNet consists of three components: a) a segmentation network which is used to obtain object masks; b) a grasp solution generator which outputs grasp solutions by fitting a minimum rectangle \cite{19} to each object based on the object mask; c) a classification network for predicting the object category label.  The operation mechanisms of the above three components are detailed as follows. 

\subsubsection{Segmentation Network} Badrinarayanan et al. \cite{20} proposed SegNet, a deep convolutional encoder-decoder architecture for real-time image segmentation. Our segmentation network is built upon this architecture.

As illustrated in Fig.~\ref{fig3}, SegNet contains two basic components: a) an encoder network which extracts visual features; b) a decoder network which classifies each pixel independently. We take 4-channel images (RGBD) as our input. In our dataset, images are center cropped to 360\(\times\)480 size from its original 540\(\times\)960 size. The final output is a binary object mask with the size of 360\(\times\)480. Specifically, the pixel value of object mask is 0 and 1 corresponding to background and object. We refer the readers to \cite{20} for the details of the SegNet.

\subsubsection{Grasp Solution Generator} In order to grasp different objects in real robotic applications, we develop a method to generate grasp solution from a object mask outputted by the segmentation network. Considering that noisy points may exist in the object mask, we only focus on the cluster that has more than 200 points. Following the approach in \cite{23} \cite{19}, we first find a convex hull for each cluster. Then we fit a minimum rectangular bounding box around each cluster based on its convex hull. From this rectangle, we calculate the rectangle center point as (\textit{x, y}) in Eq. (1) and the orientation of the rectangle relative to the horizontal axis as $\theta$ in Eq. (1).

Besides generating grasp solution, object mask can also be used to obtain a color mask. Given an original image and its object mask, we obtain the color mask as follows:
\begin{equation}
  \begin{split}
   \textit{I$_{cm}$} = \textit{I$_{m}$} \odot \textit{I$_{o}$},
  \end{split}
\end{equation}
where \textit{I$_{o}$}, \textit{I$_{m}$} and \textit{I$_{cm}$} denote the original image, object mask and color mask, respectively; and $\odot$ is  Hadamard product. As shown in Fig.~\ref{fig3}, the color mask retain the object information and eliminate the influence of illumination variations, shadow and tablecloth appearance at the same time.

The input for the following classification network (CNN)  is the object region bounded by a rectangular box  \textit{B}=(\textit{x, y, w, h}) where \textit(x, y) are the coordinates of the center of \textit{B}, and \textit{(w, h)} denote the width and height of \textit{B}. As shown in Fig.~\ref{fig3}, the box can be easily located by our object mask.  

\subsubsection{Classification Network} The last step in our grasp detection is to obtain the object category label (\textit{l} in Eq. (1)) for a complete grasp representation. In our work, we classify the object  through a CNN which has five convolutional layers and three fully-connected layers. We use a final 17-way softmax to produce a distribution over the 17 class labels.


\begin{figure*}[t]
\begin{center}
\includegraphics[width=1\linewidth]{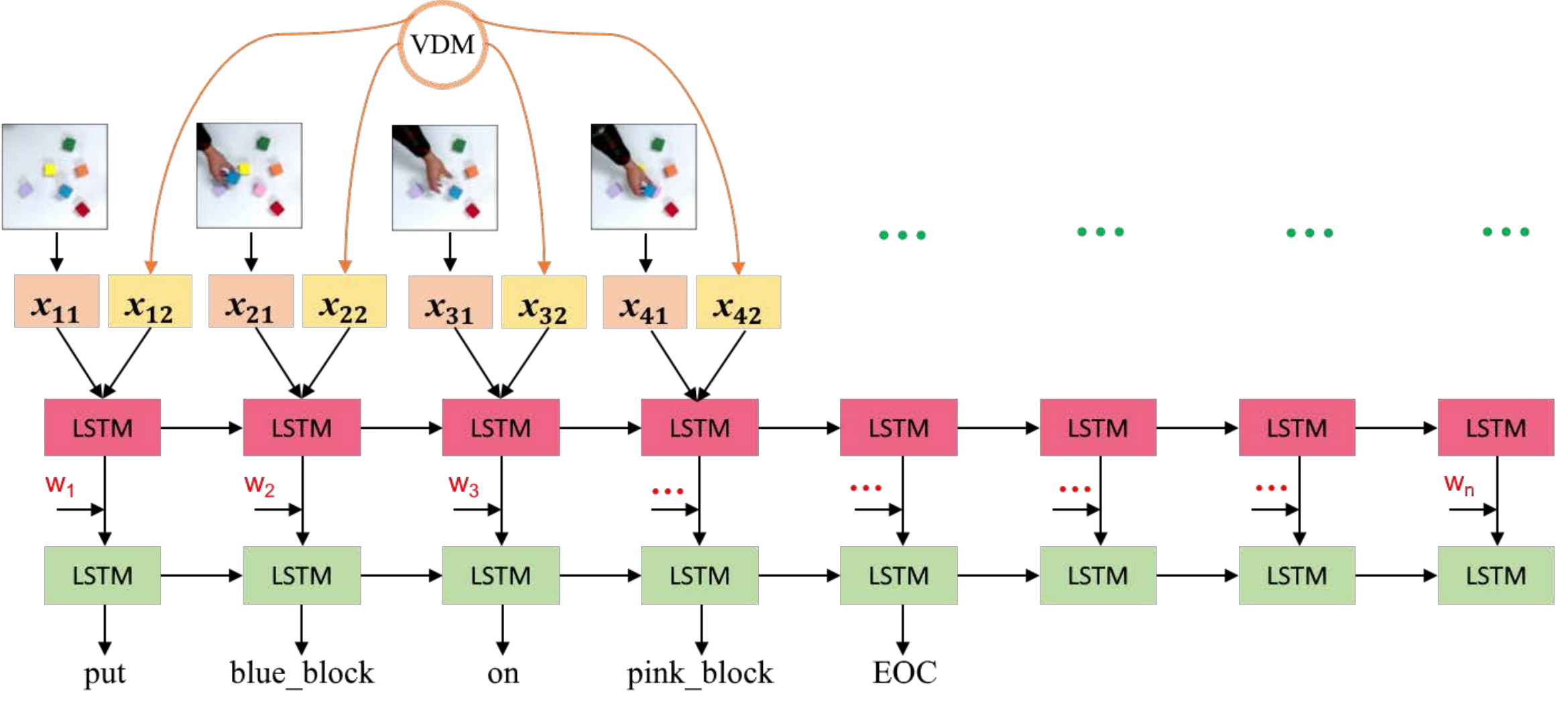}
\end{center}
\caption{An overview of our CNet. $x_{i1}$ is the feature of the $i$-th frame and $x_{i2}$ is the VDM feature. EOC means the end of the command sentence.}
\label{fig4}
\end{figure*}

In our end-to-end architecture, we use a multi-task loss \textit{L} to jointly train the segmentation network and the classification network as follows:
\begin{equation}
  \begin{split}
\textit{L} = \textit{L$_{\textit{seg}}$} + \textit{L$_{\textit{cls}}$},
  \end{split}
\end{equation}
where \textit{L$_{\textit{seg}}$} and \textit{L$_{\textit{cls}}$}  are the loss functions defined for segmentation network and classification network, respectively. Specifically, \textit{L$_{\textit{seg}}$} and \textit{L$_{\textit{cls}}$} are defined as follow:
\begin{align}
	\textit{L$_{\textit{cls}}$}(p,u) &= -\log(p_{u}),\\
	\textit{L$_{\textit{seg}}$}(m,s) &= -\frac{1}{N}\sum_{i\in Image} \log(m_{i}^{s_{i}}),
\end{align}
where \textit{p}$_{u}$ is the predicted probability of class \textit{u}, $m_{i}^{s_{i}}$ is the softmax output at pixel \textit{i} for the true label $s_{i}$, \textit{N} is the number of pixels in image. 

\subsection{CNet based Video Captioning}

As in \cite{5}, we propose to translate the human demonstration videos to command sentences by video captioning, which is implemented  with  a sequence to sequence architecture built on Long-Short Term Memory (LSTM) \cite{26}, as shown in Fig.~\ref{fig4}, called CNet in our paper. To improve the captioning performance, different from \cite{5}, our CNet inputs not only the global (frame) feature but the local (object) feature.  The details are as follows. 


\subsubsection{Problem Formulation}  Given an input video, we extract visual features \textit{X} = \{$x_{1}$, $x_{1}$,..., $x_{n}$\} from the video using a pre-trained CNN network, where $x_{i}$ is the feature of $i$th frame. The output command is presented as a sequence of word vectors \textit{S} = \{$s_{1}$, $s_{2}$,..., $s_{n}$\}, in which each vector $s_i$ represents one word vector. As aforementioned, we use a grammar-free form to describe the output command for the convenience of mapping each word in the command sentence to the real robot command.

As in \cite{venugopalan2015sequence}, we identify the most likely sentence for a given video by training a model to maximize the log likelihood of the sentence \textit{S}, given the corresponding video \textit{V} and the model parameters $\theta$.
\begin{equation}
\theta^{\hat{}} = \arg\max_{\theta} \sum_{(V,S)} \log p(S\mid V;\theta),
\end{equation}
\subsubsection{Video Features Fusion}
We improve the performance of translating videos to commands by fusing global feature and local feature. In most robotic applications, the robot manipulate objects on a fixed table and we mainly focus on the manipulated object rather than the whole scenarios. Intuitively, we can represent the changes of video contents just by the difference between the start frame and end frame, referred as video difference map (VDM). However, it is usually hard to obtain a high quality difference map due to illumination changes and shadow. To address this, our trained GNet is used to extract the color masks of start frame and end frame from a demonstration video. Then a high quality difference map can be obtained by the difference between the two color masks.

Subsequently, we extract visual features from uniformly sampled video frames $I = $\{$I_{1}$, $I_{2}$,..., $I_{n}$\} and VDM using a pre-trained CNN network, denoted as \textit{F}($I$) and \textit{F}(\textit{V}) respectively. At last, the global feature \textit{F}($I$) and local feature \textit{F}(\textit{V}) are integrated as inputs to LSTM encoder.

Since our fused features have different dimensions from the standard LSTM unit, we adjust the input unit in the first LSTM layer. In our application, for an input feature $x_{t}$ at time step \textit{t}, it  includes both $F(I)$ and $F(V)$. We only connect $F(I)$ and $F(V)$ to half of the $x_{t}$ units, since it allows different input units to specialize on modeling image features.

\subsubsection{Captioning Scheme} We use a sequence to sequence architecture to build our translation model and select LSTM as our RNN unit. 
As shown in Fig.~\ref{fig4}, in the encoding stage, the first LSTM layer takes the visual features \textit{X} = ($x_{1}$, $x_{2}$,..., $x_{n}$) as inputs and computes a sequence of hidden states \textit{H} = ($h_{1}$, $h_{2}$,..., $h_{n}$). In the decoding stage, the second LSTM layer converts the list of hidden encoder vectors \textit{${H_e}$} into the sequence of hidden decoder vectors \textit{${H_d}$}. The final list of predicted words \textit{S} is achieved by applying a softmax layer on the output \textit{${H_d}$} of the LSTM decoder layer.

\begin{figure*}[t]
\begin{center}
\includegraphics[width=1\linewidth]{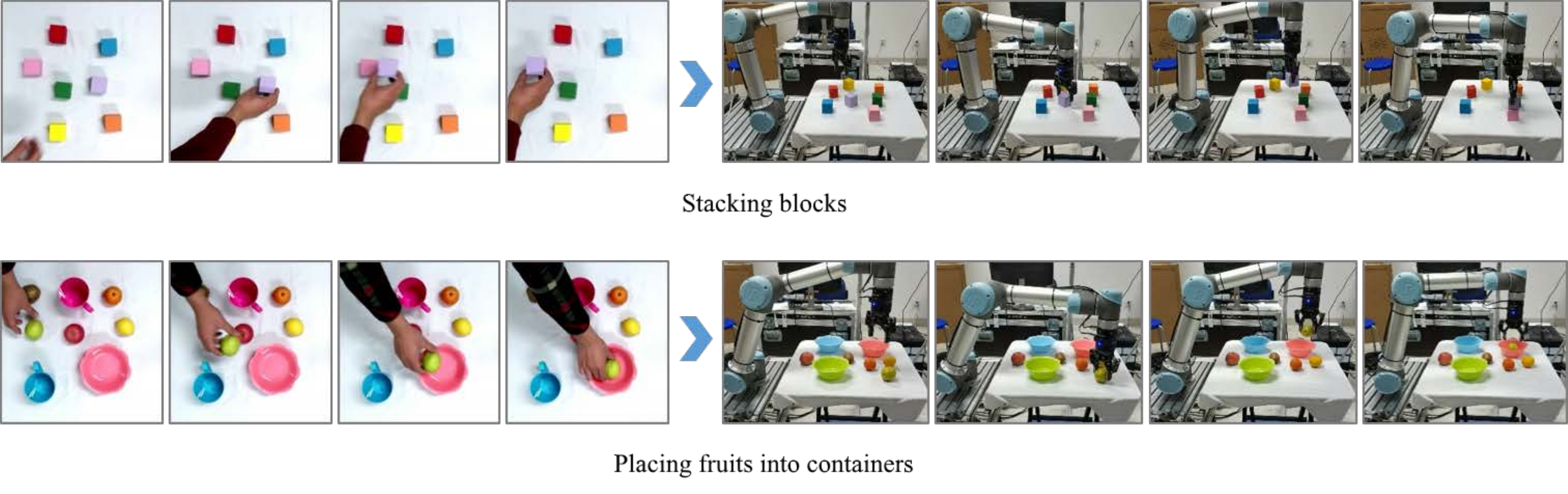}
\end{center}
\caption{Two pick-and-place tasks by UR5 robotic arm. The left part is the human demonstration and the right part is the robotic imitation.}
\label{fig7}
\end{figure*}

In the experiments, we set the number $m$ of output words equals to the number $n$ of input frames. Specifically, for the input video, we uniformly sample 30 frames in the video. Since the number $m$ of words in the output commands is always smaller than 30, we pad a special empty word to the list until we have 30 words. In practice, we add an extra word to the dictionary to denote the end of command sentences.

\section{EXPERIMENTAL RESULTS}

In this section,  we test the performance of the proposed approach by examning its two key components, GNet and CNet. For this, as illustrated in Fig.~\ref{fig7},  we design two    pick-and-place tasks, namely stacking blocks and placing fruits into containers, on a UR5 robotic arm. In what follows, we  introduce and analyze the  experimental settings and results. 


\subsection{Datasets}
Using a UR5 robotic arm, we test our approach on two tasks,  stacking blocks on a table and placing fruits into containers. For robotic study,  our scenarios are more representative and easier to reproduce compared to the 'breakfast' scenario tested in our competitor \cite{5}, which involves the tasks of  'turn off stove', 'open flour lid',   'stir egg', etc.  The videos and images for our scenarios are collected as follows. 

\subsubsection{Dataset for CNet} 



The human demonstration video is generated as follows. First, we capture raw videos that contain human actions using a Kinect camera. In each scenario, we randomly select objects and place them on a table at random locations and orientations. Then we segment raw videos into 5,980 short clips (approximately 3–6 seconds long). All short clips are center cropped to 480\(\times\)480 resolution from its original 540\(\times\)960 resolution. Each short clip is then annotated with a grammar-free command sentence that describes the human action.  We use 70\% of the dataset for training and the remaining 30\% for testing.

\subsubsection{Dataset for GNet} 


The dataset contains the images of our testing scenarios and their labels at pixel level. The RGBD images are captured with a Kinect   camera, downsampled to the size  360\(\times\)480 size from its original 540\(\times\)960 size. In total, the dataset has 5,160 images, covering  17 object categories (pink\_block, blue\_block, yellow\_block, green\_block, red\_block, purple\_block, orange\_block, pear, kiwi, apple, lemon, orange, green\_plate, blue\_plate, red\_plate, blue\_cup, red\_cup) and 35,986 annotated object regions. We use 70\% of the dataset for training and the remaining 30\% for testing.



\subsection{Experimental Settings and Results}


\subsubsection {GNet based Physical Grasping} To evaluate the grasp detection performance of our GNet, we conduct real robotic grasp experiments on a UR5 robotic arm, by comparison with three typial grasp approaches, Hybrid Grasp \cite{guo2017hybrid}, Multi-Modal Grasp \cite{kumra2017robotic} and Multi-Object Grasp \cite{34}.

\begin{figure*}[t]
\begin{center}
\includegraphics[width=0.9\linewidth]{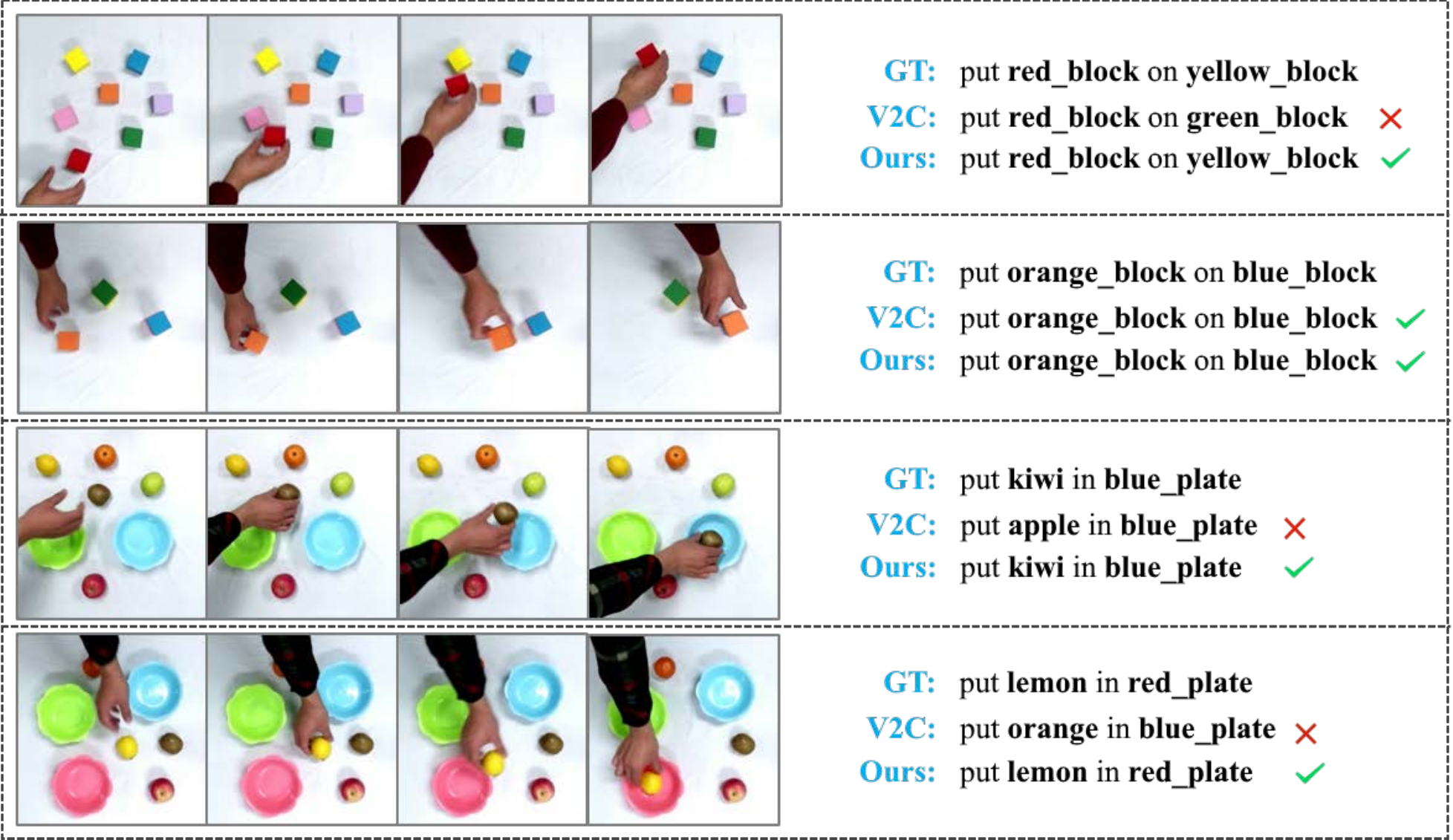}
\end{center}
\caption{Four captioning examples of the V2C \cite{5} and our CNet. For each example, the left is the human demonstration. The right shows the groundtruth caption and the captioning results yielded by V2C \cite{5} and ours.}
\label{fig8}
\end{figure*}

The experimental setups are as follows. We use two different kinds of input data to train our network. First, we use only the RGB images (GNet-RGB), then we use both the RGB and their corresponding depth images (GNet-RGBD). We train the network for 3000 epochs in an end-to-end manner using stochastic gradient descent. The batch size is empirically set to 10. Note that, \cite{kumra2017robotic} and \cite{34} took RGBD images as input, \cite{guo2017hybrid} used RGB images as input. Objects from 12 categories in our dataset were used for this experiment. For each test, several objects were randomly placed on a table at different locations and orientations. We assume all objects on the table are reachable. It is noted that we combine the 2D information yielded by GNet with the corresponding depth information to plan a robotic execution. We performed 120 trials (each object was tested 10 times) and a grasp is considered successful if the robot can grasp, raise, and hold the object for more than 3 seconds.

\begin{table}[t]
	\centering
	\caption{Comparison of grasp success rates.}
	\renewcommand\tabcolsep{10pt}
	\begin{tabular}{lc}
		\toprule  %
		Method&Grasp Success Rate\\
		\midrule
		Multi-Modal Grasp (RGBD) \cite{kumra2017robotic}&66\% (79/120)\\
		Hybrid Grasp (RGB) \cite{guo2017hybrid}&76\% (91/120)\\
		Multi-Object Grasp (RGBD) \cite{34}&90\% (108/120)\\
	    GNet-RGB&86\% (103/120)\\		
	    \textbf{GNet-RGBD}&\textbf{93\%} (\textbf{112/120})\\
		\bottomrule  %
	\end{tabular}
\end{table}

The grasp success rates are shown  Table I. It can be seen that  our GNet-RGBD reaches the success rate of 93\%, higher than other competing methods including our GNet-RGB.   This implies that the depth feature can improve grasp performance. Overall, the experiments demonstrate that our GNet behaves favorably in the real grasp task.

\subsubsection{CNet based Video Captioning} To evaluate the video captioning performance of our CNet, we baseline our approach with the V2C translation model \cite{5}. We select two popular pre-trained CNN as our feature extractor: VGG16 \cite{22}, Inception\_v3 \cite{35}. The captioning results are measured with four standard metrics, BLEU \cite{papineni2002bleu}, METEOR \cite{banerjee2005meteor}, ROUGE\_L \cite{lin2004rouge}, and CIDEr \cite{vedantam2015cider}.

The implementation is detailed below. We use 512 hidden units in LSTM. For the input, we sample 30 frames uniformly from each video.  Sequentially, we consider each command has maximum 30 words. If not enough, we pad the empty word at the end of the list until it reaches 30.  We train the CNet for 200 epochs using stochastic gradient descent. The batch size is empirically set to 15.

\begin{table}[t]
	\centering
	\caption{Comparison of captioning results. }
	\renewcommand\tabcolsep{4pt}
	\begin{tabular}{rcccc}
		\toprule  %
		&Bleu\_4&METEOR&ROUGE\_L& CIDEr\\
		\midrule
		V2C\_VGG16 &0.130&0.362&0.584&1.381\\
		V2C\_Inception\_v3 &0.136&0.365&0.593&1.437\\
		\midrule
		Ours\_VGG16&0.458&0.450&0.752&4.216\\
		Ours\_Inception\_v3&\textbf{0.510}&\textbf{0.465}&\textbf{0.775}&\textbf{4.676}\\
		\bottomrule  %
	\end{tabular}
\end{table}

The captioning results are shown in Table II. It can be observed
that our CNet achieves a better performance than V2C when we utilize VGG16 or Inception\_v3. As discussed before, the performance improvement mainly arise from the fusion of the frame feature and VDM feature.



\subsection{Demonstration of the Robotic System}


In the paper, our ultimate goal is to build an intelligent robotic system to  understand and imitate the  human actions from videos. 
For illustration, we test our approach on a UR5 robotic arm to conduct pick-and-place tasks. Specifically, we perform 20 tasks and each task is marked as success or fail depends on whether the robot can imitate the human actions correctly. The video can  be found at the following link:
\url{https://www.youtube.com/channel/UCBCRYLvS8y3cx5qXPCwXT8Q}

To test the robustness of our approach, we impose two challenges to the manipulated objects: a) They are diverse in color, shape and weight, such as the fruits and blocks shown in Fig.~\ref{fig7}; b) Their positions are randomly placed before each testing. Finally, we achieve success rates of 80\% (16/20), which obviously outperforms the success rates of V2C \cite{5}, 35\% (7/20). Fig.~\ref{fig8} shows some examples of captioning results.

From the experimental videos and results, we can see that  our approach presents stable performance across the appearance and position changing of the manipulated objects, while   our closest competitor V2C \cite{5} often fails. This verifies the robustness of our approach.  




\section{CONCLUSIONS}

In the paper, we investigated the problem of learning actions from human demonstration video based on deep neural networks. The whole grasping framework consists of two deep sub-networks: GNet and CNet. To obtain accurate command and make the video captioning more adaptive to robotic tasks, CNet was developed to fuse the global (frame) feature and the local (object) feature of interest together. GNet was designed to play two roles: provide grasp solution and generate features for the objects of interests in robotic manipulation. Extensive results in grasping tasks demonstrated the superiority of the proposed system over state-of-the-art work.

\section*{ACKNOWLEDGMENT}

This work was supported by the National Key Research and
Development Plan of China under Grant 2017YFB1300205,
NSFC Grant no. 61573222, and Major Research Program of
Shandong Province 2018CXGC1503.



\bibliographystyle{ieeetr}
\bibliography{egbib}
\end{document}